\documentclass{article}

\usepackage{arxiv}

\usepackage[utf8]{inputenc} 
\usepackage[T1]{fontenc}    
\usepackage{hyperref}       
\usepackage{url}            
\usepackage{booktabs}       
\usepackage{amsfonts}       
\usepackage{nicefrac}       
\usepackage{microtype}      
\usepackage{lipsum}		
\usepackage{graphicx}
\usepackage{doi}
\usepackage{amsmath}
\usepackage{float}
\usepackage[title]{appendix}
\usepackage{tabularx}
\usepackage{caption}
\usepackage{subcaption}
\usepackage{todonotes}
\usepackage{hyperref}
\usepackage[flushleft]{threeparttable}
\hypersetup{colorlinks,allcolors=black}

\newcommand{\mat}[1]{\mathbf{#1}}           
\newcommand{\R}{\mathbb{R}}

\title{A template for the \emph{arxiv} style}

\usepackage[citestyle=numeric,sorting=none]{biblatex}
\title{SimpleTRON: Simple Transformer with O(N) Complexity}

\author{ \href{https://orcid.org/0000-0000-0000-0000}{\includegraphics[scale=0.06]{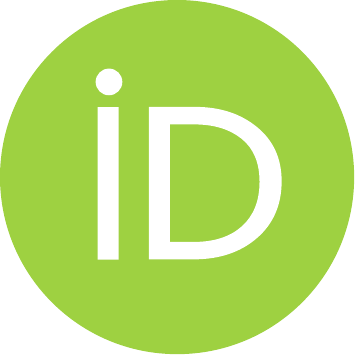}\hspace{1mm}Uladzislau ~Yorsh}\\
	Faculty of Information Technology\\
	Czech Technical University in Prague\\
	Prague, Czech Republic\\
	\texttt{yorshula@fit.cvut.cz} \\
    \And
	\href{https://orcid.org/0000-0002-7194-1874}{\includegraphics[scale=0.06]{orcid.pdf}\hspace{1mm}Alexander ~Kovalenko}\\
	Faculty of Information Technology\\
	Czech Technical University in Prague\\
	Prague, Czech Republic\\
	\texttt{kovalale@fit.cvut.cz} \\
	\And
	\href{https://orcid.org/0000-0000-0000-0000}{\includegraphics[scale=0.06]{orcid.pdf}\hspace{1mm}Vojtěch ~Vančura}\\
	Faculty of Information Technology\\
	Czech Technical University in Prague\\
	Prague, Czech Republic\\
	\texttt{vancurv@fit.cvut.cz} \\
	\And
	\href{https://orcid.org/0000-0003-0616-4340}{\includegraphics[scale=0.06]{orcid.pdf}\hspace{1mm}Daniel ~Vašata}\\
	Faculty of Information Technology\\
	Czech Technical University in Prague\\
	Prague, Czech Republic\\
	\texttt{daniel.vasata@fit.cvut.cz} \\
	\And
	\href{https://orcid.org/0000-0003-1433-0089}{\includegraphics[scale=0.06]{orcid.pdf}\hspace{1mm}Pavel ~Kordík}\\
	Faculty of Information Technology\\
	Czech Technical University in Prague\\
	Prague, Czech Republic\\
	\texttt{kordik@fit.cvut.cz} \\
	\And
	\href{https://orcid.org/0000-0002-6938-5426}{\includegraphics[scale=0.06]{orcid.pdf}\hspace{1mm}Tomáš  ~Mikolov}\\
	Czech Institute of Informatics,\\ Robotics and Cybernetics\\
	Czech Technical University in Prague\\
	Prague, Czech Republic\\
	\texttt{tomas.mikolov@cvut.cz} \\
}

\begin{document}

\newcolumntype{L}[1]{>{\raggedright\arraybackslash}p{#1}}
\newcolumntype{C}[1]{>{\centering\arraybackslash}p{#1}}
\newcolumntype{R}[1]{>{\raggedleft\arraybackslash}p{#1}}

\maketitle

\begin{abstract}
  
  In this paper, we propose that the dot product pairwise matching attention layer, which is widely used in Transformer-based models, is redundant for the model performance. Attention, in its original formulation, has to be seen rather as a human-level tool to explore and/or visualize relevancy scores in sequential data. However, the way how it is constructed leads to a significant computational complexity. Instead, we present SimpleTRON: Simple Transformer with O(N) Complexity, a simple and fast alternative without any approximation that, to the best of our knowledge, outperforms existing sub-quadratic attention approximation models on several tasks from the Long-Range Arena benchmark. Moreover, unlike other approximation models, SimpleTRON does not have any architecture-related overhead therefore can be seen as a purely linear Transformer-like model.
  
\end{abstract}

\section{Introduction}

Initially designed for natural language processing, the Transformer architecture \cite{vaswani2017attention} emerged in a wide range of other domains and quickly became a state-of-the-art in language modeling \cite{devlin2019bert} as well as in generative tasks \cite{radford2019language,brown2020language}, image processing \cite{dosovitskiy2021image,touvron2021training}, speech recognition \cite{shi2020emformer}, reinforcement learning \cite{chen2021decision}, and others. 
From the original paper named "Attention is All You Need" \cite{vaswani2017attention} on, it seems to be widely considered that the query-key-value framework, which implies a global pairwise comparison between query and key tokens, is a necessary condition for the model performance. Even though such a mechanism allows a human-comprehensible visualization of interactions between the tokens, unveiling the interpretability up to some extent, an element-wise token comparison leads to a quadratic complexity both in terms of time and space. Therefore, even though the original Transformer architecture virtually can handle arbitrarily long range dependencies given the infinite compute, which is in opposite to most recurrent neural networks \cite{sutskever2013training}, the complexity of regular full-rank attention limits Transformer applications when long sequences are required.

In this paper we present a SimpleTRON model with a SimpleAttention mechanism as an extremely simple yet efficient solution to replace the original quadratic complexity softmax attention. The proposed mechanism not only possesses linear time and memory complexity, but outperforms the current state-of-the-art Transformers on the text classification, matching and ListOps tasks from the LRA \cite{tay2020long} benchmark, which became a widely applied test for sequence processing models. Moreover, since the SimpleAttention has analogous building blocks as the original attention it is suitable for transfer learning as one can use pre-trained weights from the existing transformer models.

\section{Related Work}

As a restrictive limitation, the computational complexity of the original model motivated the community to quest for the solution in order to approximate the architecture with asymptotically faster models \cite{tay2020efficient}. Thus, recently, a dizzying number of so-called "Efficient Transformers" appeared. Each of these implementations applied some notion of sparsity to the otherwise dense attention mechanism and reached a sub-quadratic complexity with comparable performance.

Among the solutions to rationalize the Transformer complexity, there were engineering approaches such as sparse attention \cite{roy2020efficient, child2019sparse}, graph attention \cite{ahmad2021gate} or compressive attention \cite{rae2019compressive} that maps past hidden activations to a smaller set of compressed representations, allowed to use longer sequences at comparable compute. Further engineering methods include Longformer \cite{beltagy2020longformer}, where attention mechanism is a combination of a windowed local-context self-attention and a global attention that encodes inductive bias, Coformer \cite{gulati2020conformer} and Attention Augmented CNN \cite{bello2020attention} which are hybrid architectures of CNN augmented Transformers, Imputer \cite{chan2020imputer} -- the model that generates output sequences iteratively via imputations and dynamic programming, Reformer \cite{kitaev2020reformer} using dot-product attention and reversible residual layers, N-gram Masked Self-Attention \cite{chelba2020faster} etc.

Another branch of research to reduce Transformer complexity is dedicated to matrix and kernel approximations based on strong mathematical basis. That includes Performer \cite{choromanski2021rethinking}, which uses kernel approximation, factorized attention \cite{shen2020efficient}, random feature attention \cite{peng2021random} or, for example, Nystr\"omformer \cite{xiong2021nystromformer} using Nystr\"om matrix approximation. Finally, learnable kernel approximation was presented by Chowdhury et al. \cite{chowdhury2021learning}, where the authors reported trainable kernel by learning the spectral distribution and approximation of the Transformer kernel as a dot product between spectral feature maps.

Additionally there are several studies that changed the whole concept of attention and replaced it with Fast Fourier Transform (FFT) which does not require any training \cite{leethorp2021fnet}, or used dense layers to mix the tokens along both axes \cite{tolstikhin2021mlpmixer}. Another work \cite{anonymous2022patches} reported vision transformer-inspired\cite{dosovitskiy2021image} model  that independently mixes the spatial
and channel locations of image patches using depth-wise convolutions, outperforming existing Transformer-based solutions for image recognition.


\section{The Model}

The original multi-head attention layer utilizes the softmax-normalization of a head-wise product of query and transposed key matrices combined with the value matrix as:
\[ 
    \text{Attention}(\mat{Q}_h,\mat K_h, \mat V_h) = \text{softmax}\left(\frac{\mat Q_h \mat K_h^T}{\sqrt{d}}\right)\mat V_h, 
\]
where $\mat Q_h, \mat K_h, \text{and } \mat V_h \in \R^{L, d}$ are the query, key, and value, respectively, corresponding to the $h$-th head, $d$ is the query dimensionality, and $L$ is the length of the input sequence. The head-wise inputs to the operation are obtained by splitting the $\mat Q, \mat K, \mat V \in \R^{L, D_{\text{hid}}}$ matrices across the hidden dimension axis $D_{\text{hid}}$ into $N_h$ pieces of size $d = D_{\text{hid}} / N_h$ corresponding to $N_h$ heads. 
The input $X \in \R^{L, D_{\text{E}}}$ is transformed to the $\mat Q, \mat K, \text{and } \mat V$ matrices by a linear transformation using matrices $\mat Q^*, \mat K^*, \mat V^* \in \R^{D_{\text{E}}, D_{\text{hid}}}$ and biases $\mat q^*, \mat k^*,  \mat v^*  \in \R^{L, D_{\text{hid}}}$ as parameters:
\[ 
    \mat{Q} = \mat X \mat Q^* + \mat q^*,\quad \mat{K} = \mat X \mat K^* + \mat k^*,\quad \mat{V} = \mat X \mat V^* + \mat v^*.
\]
The final output of the attention layer is then produced by applying another linear layer on a concatenation of all heads and adding the duplicated input $X$ that corresponds to a skip connection:
\[ 
    \text{SelfAttention}(\mat X) = \mat X + \big(\text{Attention}(\mat{Q}_1,\mat K_1, \mat V_1), \dotsc,  \text{Attention}(\mat{Q}_{N_h},\mat K_{N_h}, \mat V_{N_h})\big)\mat W + \mat w,
\]
where $\mat W \in \R^{D_{\text{hid}}, D_{\text{E}}}$ and $\mat w \in \R^{L, D_{\text{E}}}$ are the parameters of the linear layer.

The $\mat Q_h, \mat K_h, \text{and } \mat V_h$ are rectangular matrices with the first dimension typically dominating the second one. Thus, the quadratic complexity appears upon the $\mat Q_h \mat K_h^T$ operation with respect to the sequence length. Swapping matrix multiplication order (first $\mat K_h^T \mat V_h$, then multiply with $\mat Q_h$) would reduce complexity to linear. However, softmax non-linearity forbids such shuffling. Here, we applied several major tweaks to the model in order to reach linear complexity and to improve performance:
\begin{itemize}
    \item Reject the softmax nonliearity.
    \item Change the order of matrix multiplication to avoid quadratic complexity.
    \item (Optionally) Remove the linear layer producing the final output of the attention layer.
\end{itemize}

Therefore we obtain a no-softmax attention with the direct q-k-v product, which could be described by the following simple formula for an attention operation on a single head:
\begin{equation*}
\begin{alignedat}{3}
    \text{SimpleAttention}(\mat{Q}_h,\mat K_h, \mat V_h) & = \frac{1}{\sqrt{L}}\mat Q_h \big(\mat K_h^T \mat V_h\big).
\end{alignedat}
\end{equation*}
When the linear layer is not used to produced the final output, then the q-k-v product concatenated for all heads goes directly to the residual sum with duplicated input from skip connection.

Unlike the linear mixing models such as \cite{leethorp2021fnet}, this transformation is not linear in terms of input $\mat X$.
To see this just note that 
\[
    \mat Q_h = \mat Q \mat I_h = \mat X \mat Q^*  \mat I_h + \mat q^* \mat I_h
\]
and analogously for $\mat K_h$ and $\mat V_h$, where $I_h \in \R^{D_{\text{hid}}, d}$ is a matrix with all entries zero except identity matrix $d\times d$ on rows from $d(h-1) + 1$ to $d h$, i.e. the multiplication $\mat Q \mat I_h$ takes exactly those columns from matrix $\mat Q$ that correspond to the $h$-th head.
Hence, we obtain
\[
    \text{SimpleAttention}(\mat X) = \frac{1}{\sqrt{L}}(\mat X \mat Q^*  \mat I_h + \mat q^* \mat I_h)\big((\mat X \mat K^*  \mat I_h + \mat k^* \mat I_h)^T(\mat X \mat V^*  \mat I_h + \mat v^* \mat I_h)\big).
\]
This equation resembles the quadratic form multiplied again by the input.

\begin{figure}
\centerline{\includegraphics[width=16cm]{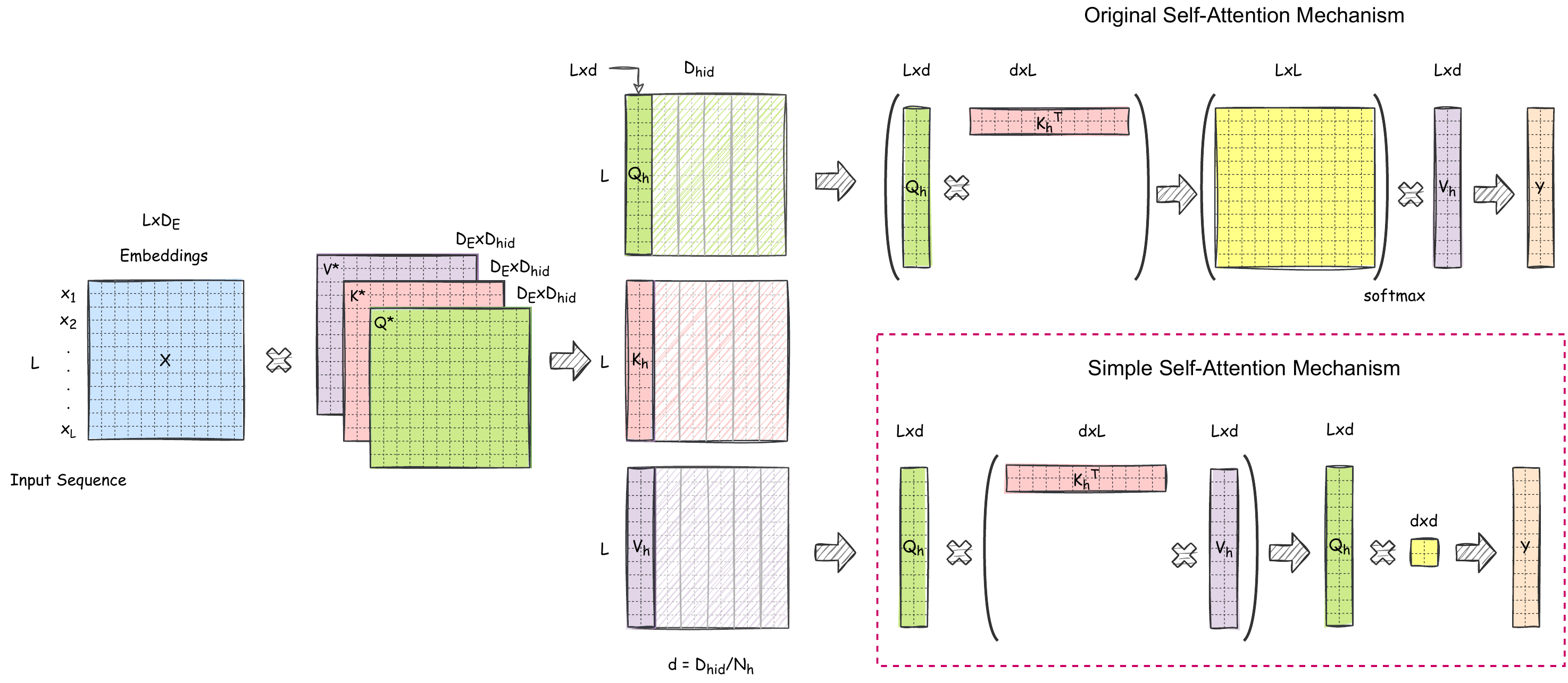}}
\caption{Illustration of the Self-Attention calculation in the vanilla Transformer (top) and \emph{SimpleTRON} Self-Attention (bottom) dashed blocks representing matrices: $\mat X$ - input sequence $\{ x_1, x_2... , x_n\}$ of the length $L$, $\mat Q^*$ - query, $\mat K^*$ - key, $\mat V^*$ - value square matrices of dimensions $D_{\text{E}}\times D_{\text{hid}}$, $d$ is a dimension of a single head ($d = D_{\text{hid}}/N_h$), $D_E$ is an embedding dimensionality.}
\label{fig:simple attention}
\end{figure}

We further refer the above-mentioned mechanism as \emph{SimpleAttention}, and the model as \emph{SimpleTRON} which stands for Simple Transformer with O(N) Complexity. The matrix operations within a single head of \emph{SimpleAttention} is illustrated in  Figure \ref{fig:simple attention}.

\section{Experiments}

\subsection{LRA benchmark}

Even though numerous sub-quadratic complexity approximations of the vanilla Transformer claimed comparable or even superior performance to the original model, it is fair to express that each model can be task-dependent and possess strikingly different results upon modality. Moreover, some benchmark test can be parameter-dependent, thus bigger models can perform better. Therefore up to some point effective evaluation of the Transformer-like models was uncertain, due to the absence of a unified and systematic benchmark. In this regard, Tay et al. \cite{tay2020long} published the benchmark for efficient Transformer models called "Long Range Arena" (LRA), that consists of task of various data types and modalities, where data is presented in sequences ranging from 1K to 16K tokens.

We use LRA \cite{tay2020long} as the standardised benchmark for efficient Transformer evaluation:

\begin{itemize}
    \item Following the recommendations from \cite{tay2020long}, we replicate the learning schedule and all the hyperparameters that relate to our model architecture, while keeping additional parametrization below $10\%$.
    \item To reproduce the experimental setup from \cite{tay2020long}, we used the gradient accumulation in order to simulate larger batch sizes.
    \item Given the stochastic weight initialization and sampling, each model was trained for 5 times to observe model behavior, accuracy variance and to avoid so-called \textit{black swans} -- random seeds that give radically different results \cite{picard2021torchmanualseed3407}. Best results are reported in Table \ref{tab:baseline}.
    \item As we focus on NLP domain in the present work, we test out model on three LRA tasks -- BPE text classification, information retrieval and ListOps.
    \item Since our models tend to converge slower in terms of number of iterations, we prolonged the training on matching and ListOps tasks to $15$ k steps.
\end{itemize}


%
    

The model was implemented in PyTorch library \cite{NEURIPS2019_9015}.

\subsection{\textit{SimpleTRON} transfer learning}

In the beginning of the paper, we raised a question, whether we need a pairwise matching attention layer or any of its approximations. To find the answer we performed a simple experiment: 

First, the \emph{SimpleTRON} model was trained regularly, reaching its top accuracy.
As q-k-v matrices are of the same dimensions in both \emph{SimpleAttention} and original \emph{SoftmaxAttention}, thus the weights are interchangeable. Therefore, we transferred the trained weights from \emph{SimpleAttention} to \emph{SoftmaxAttention}, froze q-k-v layers and retrained the rest of the model. The logic behind such experiment was quite simple: if we need a pairwise comparison in q-k product, then the model won't be able to reach the efficiency of a vanilla Transformer as q-k-v layers are frozen and not trained optimally.

Moreover, as we know the community performed an immense effort to pretrain large language models on comprehensive datasets, allowing many researchers and companies reap all the benefits of transfer learning by fine-tuning pretrained models on specific tasks from various domains \cite{sun2019fine, zhang2021revisiting, GPT-2_2021}. Lately it was proposed, that learning abilities of the Transformer models trained on a extensive language dataset can overcome NLP modality and used as a universal computational engines \cite{lu2021pretrained}. On the other hand, such training is an extremely resource demanding process \cite{dale2021gpt} with a considerable carbon footprint \cite{patterson2021carbon}. Therefore, simply of of curiosity, we tried to apply fine tuning for text classification on our \emph{SimpleTRON} model using weight from pretrained BERT \cite{devlin2019bert}. This operation of weight transfer is applicable to \emph{SimpleTRON} architecture as the size and dimensionality of the model layers can be fully identical and, therefore, transferable.

\newcommand{\tableonewidth}{1.75cm}

\begin{table}
    \centering
    \begin{tabular}{ l | C{2.5cm} | C{\tableonewidth} | C{\tableonewidth} | C{\tableonewidth} } 
         \hline
         & & & & \\
         Model & Complexity &  Classification & Matching & ListOps \\ [0.5ex] 
         & & & & \\
         \hline\hline
          & & & & \\
         Random & $\mathcal{O}(1)$ & 50.00 & 50.00 & 10.00\\
         Transformer & $\mathcal{O}(L^2)$ & 64.27 & 57.46 & 36.37 \\
          & & & &\\
         \hline
          & & & &\\
         Synthesizer & $\mathcal{O}(L^2)$ & 61.68 & 54.67 & 36.99\\ 
         Sinkhorn Trans. & $\mathcal{O}(B^2 + (N/B)^2)$ & 61.20 & 53.83 & 33.67 \\
         Sparse Trans. & $\mathcal{O}(L\sqrt{L})$ & 63.58 & 59.59 & 17.07\\
         Reformer & $\mathcal{O}(L\log{L})$ & 56.10 & 53.40 & 37.27\\
         Local Attention & $\mathcal{O}(LK)$ & 52.98 & 53.39 & 15.82\\
         Longformer & $\mathcal{O}(LK)$ & 62.85 & 56.89 & 35.63\\
         Linformer & $\mathcal{O}(L)$ & 53.94 & 52.27 & 35.70 \\
         BigBird & $\mathcal{O}(L)$ & 64.02 & 59.29 & 36.05\\
         Linear ELU & $\mathcal{O}(L)$ & 65.90 & 53.09 & 16.13\\
         Performer & $\mathcal{O}(L)$ & 65.40 & 53.82 & 18.01\\
          & & & &\\
         \hline
          & & & &\\
         GMM-RKS & $\mathcal{O}(L)$ & 66.20 & 58.74 & 18.15\\
         FastFood-RKS & $\mathcal{O}(L)$ & 65.91 & 57.47 & 18.20\\
         Generative-RKS & $\mathcal{O}(L)$ & 66.37 & 59.02 & 17.80\\
         GMM-PRF & $\mathcal{O}(L)$ & 62.70 & 59.64 & 36.95\\
         FastFood-PRF & $\mathcal{O}(L)$ & 64.69 & 67.90 & 37.25\\
         Generative-PRF & $\mathcal{O}(L)$ & 62.39 & 67.18 & 37.10\\
          & & & & \\
         \hline
          & & & & \\
         \textbf{Simple (ours)} & $\mathcal{O}(L)$ & \textbf{66.75} & \textbf{73.92} & \textbf{37.45}\\
         \textbf{Simple-Res (ours)} & $\mathcal{O}(L)$ & \textbf{66.65} & \textbf{74.83} & \textbf{37.10}\\
         \textbf{Simple-ResL (ours)} & $\mathcal{O}(L)$ & \textbf{66.71} & \textbf{73.59} & \textbf{37.55}\\
          & & & &\\
         \hline
    \end{tabular}
    \caption{Baseline and proposed models on the three LRA tasks. We denote sequence length as $L$, attention span as $K$ and Sinkhorn model block size as $B$. The notation for our models is: \textbf{Simple} - \emph{SimpleAttention} without both skip connection and linear layer, \textbf{Simple-Res} - \emph{SimpleAttention} with skip connection and without linear layer, and \textbf{Simple-ResL} - \emph{SimpleAttention} with skip connection and linear layer behind the q-k-v multiplication.}
    \label{tab:baseline}
\end{table}

\section{Results and Discussion}

\subsection{LRA}

\paragraph{Number of parameters}

By removing a linear layer following the q-k-v product, our model in fact had less parameters than its counterparts given the restrictions reported in \cite{tay2020long}. Which is only a small margin, however worth mentioning as the authors placed parametrization restrictions in their paper.

\paragraph{Training speed}

By swapping the q-k-v product matrices and avoiding any kind of approximation we've reached a truly linear complexity with a respect to the input length. It has to be emphasized, that the most of linear attention approximations reporting linear complexity, in fact omitting high architecture-dependent multiplier, which should be taken into account in practice.

\paragraph{Memory efficiency}

The current model, given the above-mentioned LRA tasks, was found to be an order of magnitude more memory efficient in comparison with the vanilla transformer. 

\paragraph{Classification accuracy}

To observe the model behaviour and accuracy variance, we train our models for 5 times to avoid the so-called \textit{black swans} -- random seeds that give radically different results \cite{picard2021torchmanualseed3407}. As a result, our model had shown $66.75 / 73.92 / 37.45 \%$ top accuracy on the test classification/matching/ListOps splits and $66.61/73.74/37.15 \%$ mean test accuracy respectively, outperforming other known linear approximation of attention mechanism.

\paragraph{Normalization} The original Transformer model uses the $1/\sqrt{d}$ normalization term of the $\mat Q_h \mat K_h^T$ product to counteract the vector magnitude explosion and the following decreased gradient flow through the Softmax. Since we don't use any saturating function in the attention module, our model works without any normalization terms. However, we have found the $1/\sqrt{L}$ term to be useful for a more stable convergence.

\paragraph{Convergence}

\begin{figure}[h]
     \centering
     \begin{subfigure}[b]{0.33\textwidth}
         \centering
         \includegraphics[width=\textwidth]{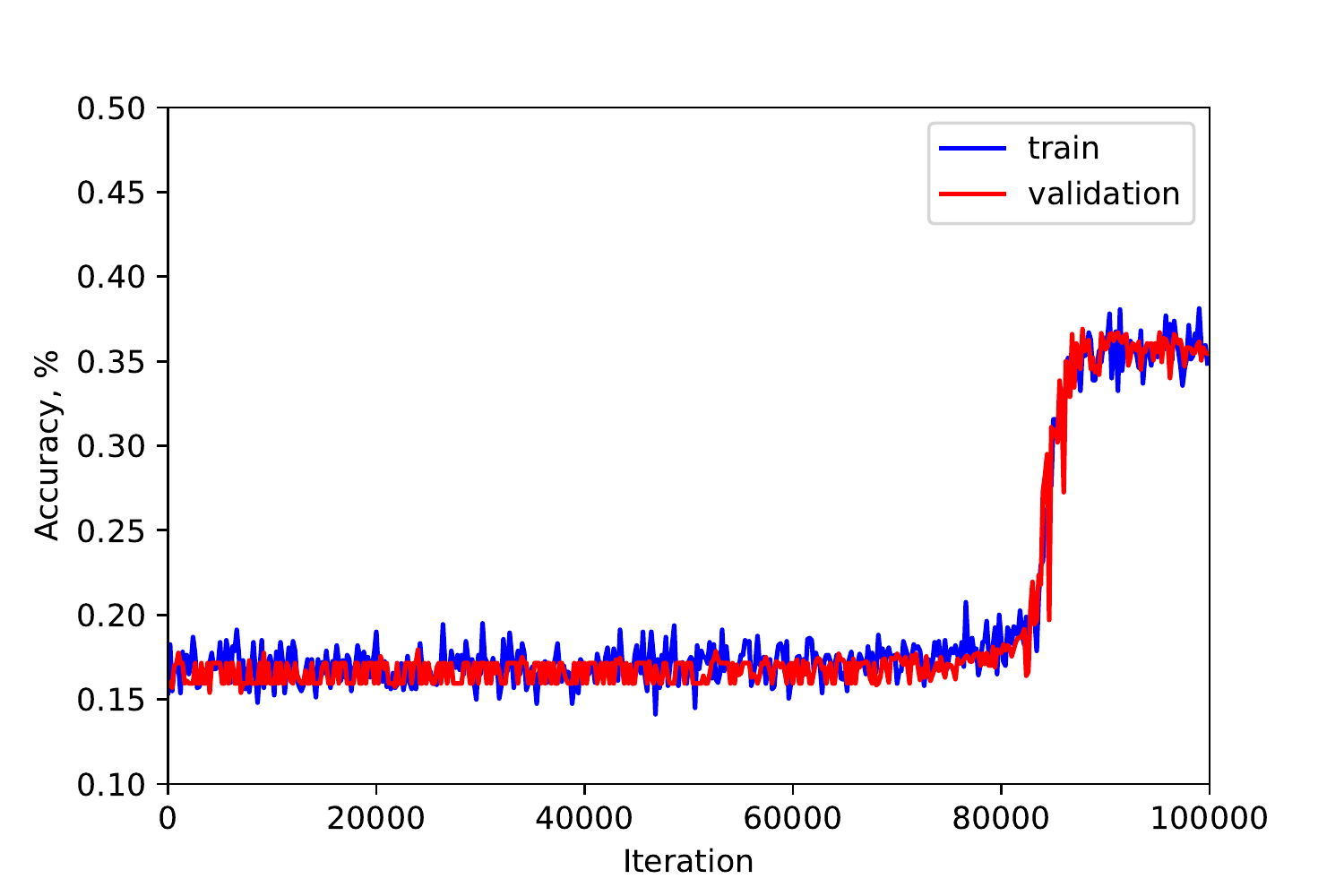}
         \caption{ListOps}
         \label{fig:lops acc}
     \end{subfigure}
     \hfill
     \begin{subfigure}[b]{0.33\textwidth}
         \centering
         \includegraphics[width=\textwidth]{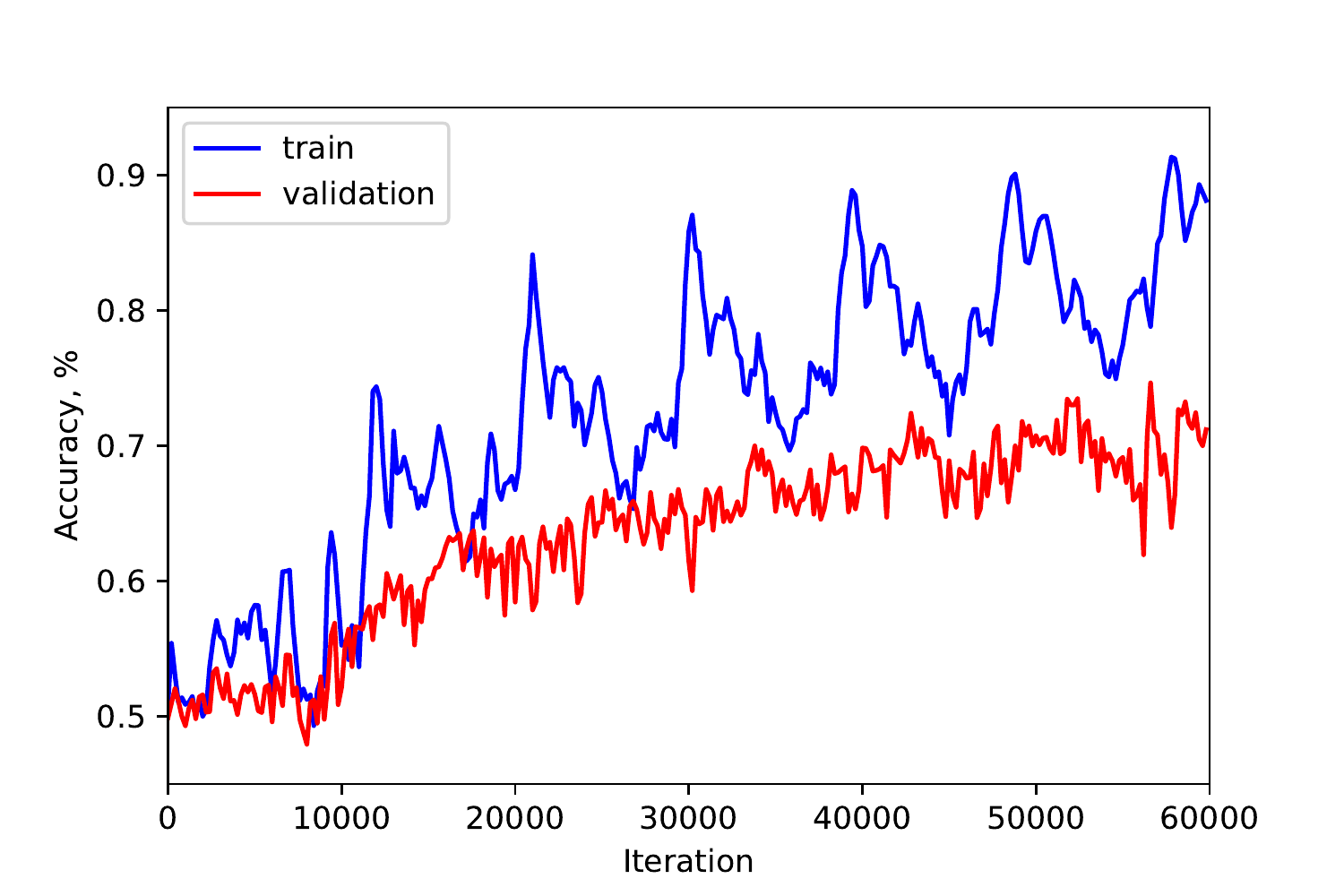}
         \caption{Matching}
         \label{fig:matching acc}
     \end{subfigure}
     \begin{subfigure}[b]{0.33\textwidth}
         \centering
         \includegraphics[width=\textwidth]{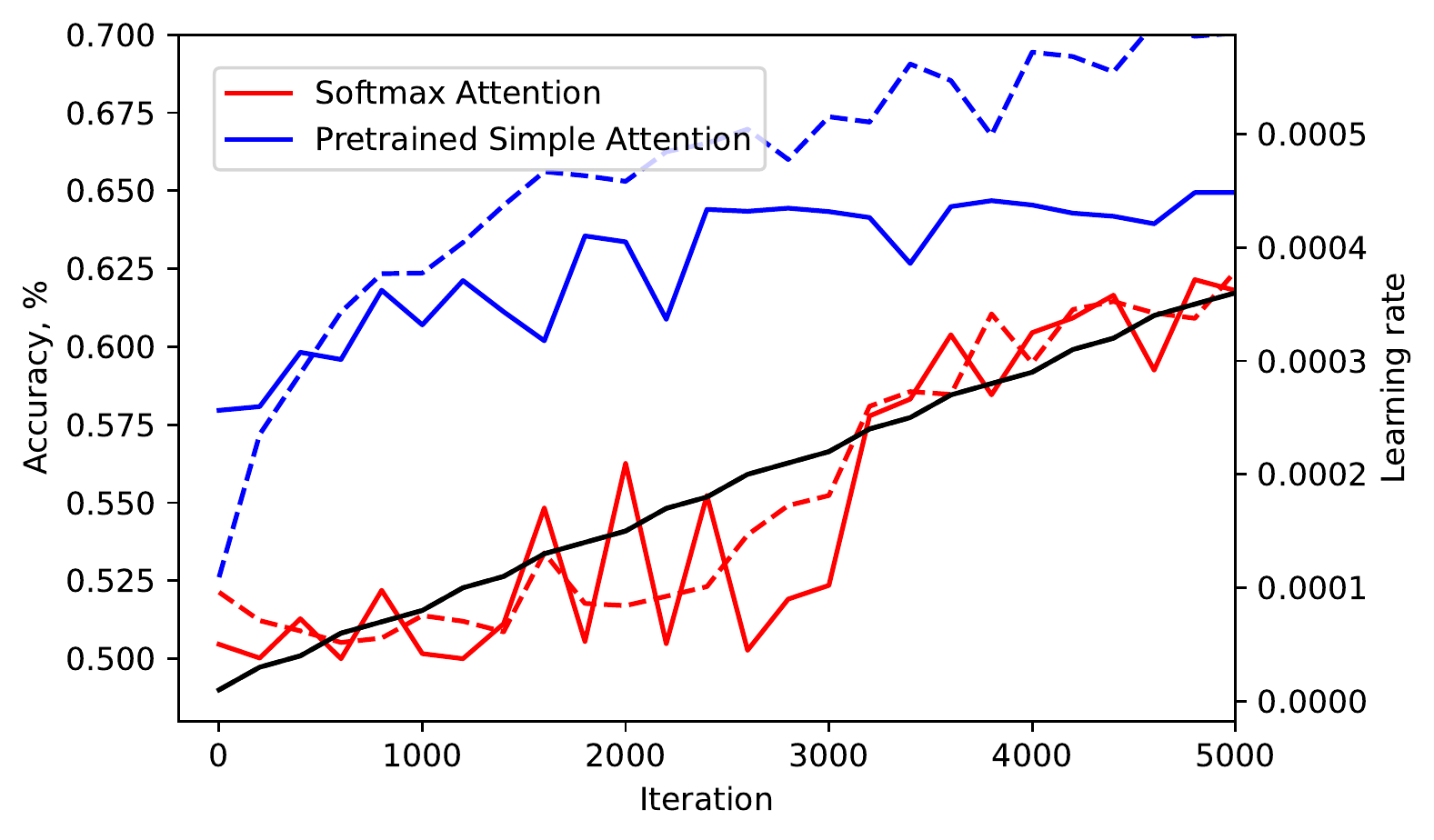}
         \caption{Matching}
         \label{fig:transfer train}         
     \end{subfigure}
    \caption{Training evolution plots of example runs on ListOps \ref{fig:lops acc}, Matching \ref{fig:matching acc}. \ref{fig:transfer train} Training (dashed lines) and validation (solid lines) curves of the original vanilla Transformer (red lines) architecture and transformer with frozen q-k-v layers (blue lines) transferred from \emph{SimpleAttention} model.}
    \label{fig:lops and matching}
\end{figure}

We have found our model to be converging slower than competitors in terms of number of iterations, but this is being counteracted by faster computation. Furthermore, we argue that according to the validation loss and accuracy behavior on some tasks better results could be obtained by further training of our model, see Figure \ref{fig:lops acc} and \ref{fig:matching acc}.

\paragraph{Compressed representation} The $\mat Q_h \mat K_h^T \mat V_h$ product in \emph{SimpleAttention} may be interpreted as a comparison of an uncompressed input projection $\mat Q_h$ with its compressed representation $\mat K_h^T \mat V_h$.

\paragraph{Attention transfer}

Transferred weights from, \emph{SimpleAttention} model to the original \emph{SoftmaxAttention} model has shown an interesting behaviour: having about 30\% less trainable parameters and in fact no ability to learn pairwise relation between the tokens in the sequences, the model trained up to the original accuracy (Figure \ref{fig:transfer train}) of the vanilla transformer, however in much less of training epochs. We assume, that given the frozen gradients in q-k-v layers, such \emph{SimpleAttention} pre-training is an evidence of the pairwise token comparison redundancy. This also opens a way of fast (re)training of already deployed models.

\subsection{Linear Layer Elimination and Excessive Depth}

It has to be emphasized, that the models, where the order of matrix multiplication in the attention head does is simply swapped without any further modification often do not converge. Successful training is possible only in the case, when the number of blocks is low (i.e. 4 blocks for the text classification task). Even though, at the very early stage of training, deeper models with simple attention are on-par with the vanilla Transformer and after certain number of epochs work no better than a random choice.

One of the pathways in order to reach a stable and efficient training, is to remove the linear layer that follows output of the attention as described above. However, it was empirically discovered for larger models with a higher number of parameters that are usually applied in practice (i.e. comparable with BERT \cite{devlin2019bert} by a number of parameters) the original \emph{SimpleTRON} architecture shows performance lagging behind the vanilla Transformer. Moreover, in some cases, a linear layer following attention operation is technically necessary when the dimensionality of $D_{\text{E}}$ differs from $D_{\text{hid}}$. Another case, when the linear layer would be beneficial is a weight transfer from pretrained transformer like model. 

\begin{figure}
\centerline{\includegraphics[width=4cm]{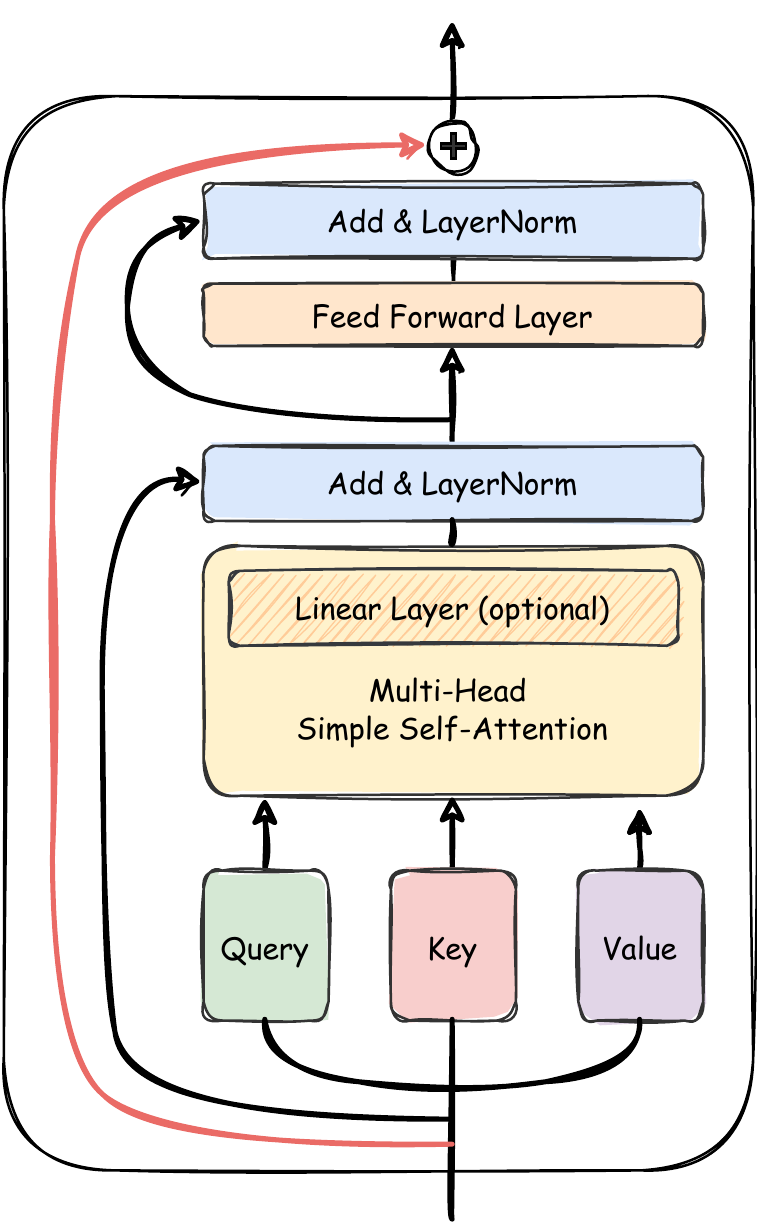}}
\caption{Illustration of the \emph{SimpleAttention} block with a double skip connection.}
\label{fig:simple block}
\end{figure}

Another option is adding skip-connections \cite{he2016deep} through the \emph{SimpleTRON} block. Nevertheless, the original Transformer block already has skip connection from the block input to \emph{Layer Normalization} layer, in the present implementation we added another skip connection (shown in red on Figure \ref{fig:simple block}). Therefore, we were able to train larger models of arbitrary depth. In this case, the presence of an additional linear layer does not have any deleterious effect on the model's convergence.

The reason for the deeper models to fail on training is that the weights of \emph{SimpleAttention} output tend to be symmetrical in the deeper blocks of the model. Additional skip connections lead to higher variance in weights and therefore better inference ability of the model (see Figure \ref{fig:attention weight std} in the Appendix). Furthermore, to show that the models with skip connections perform well on the LRA dataset, we performed the experiments using the models with skip connections with or without the linear layer. The result are consistent with ones of the original \emph{SimpleTRON} model, while model the linear layer usually do not converge without skip connections (see Table \ref{tab:baseline}). It is worth mentioning, that we could not obtain any accuracy gain by stacking more \emph{SimpleTRON} together neither with nor without skip connections.


\begin{figure}[h]
     \centering
     \begin{subfigure}[b]{0.33\textwidth}
         \centering
         \includegraphics[width=\textwidth]{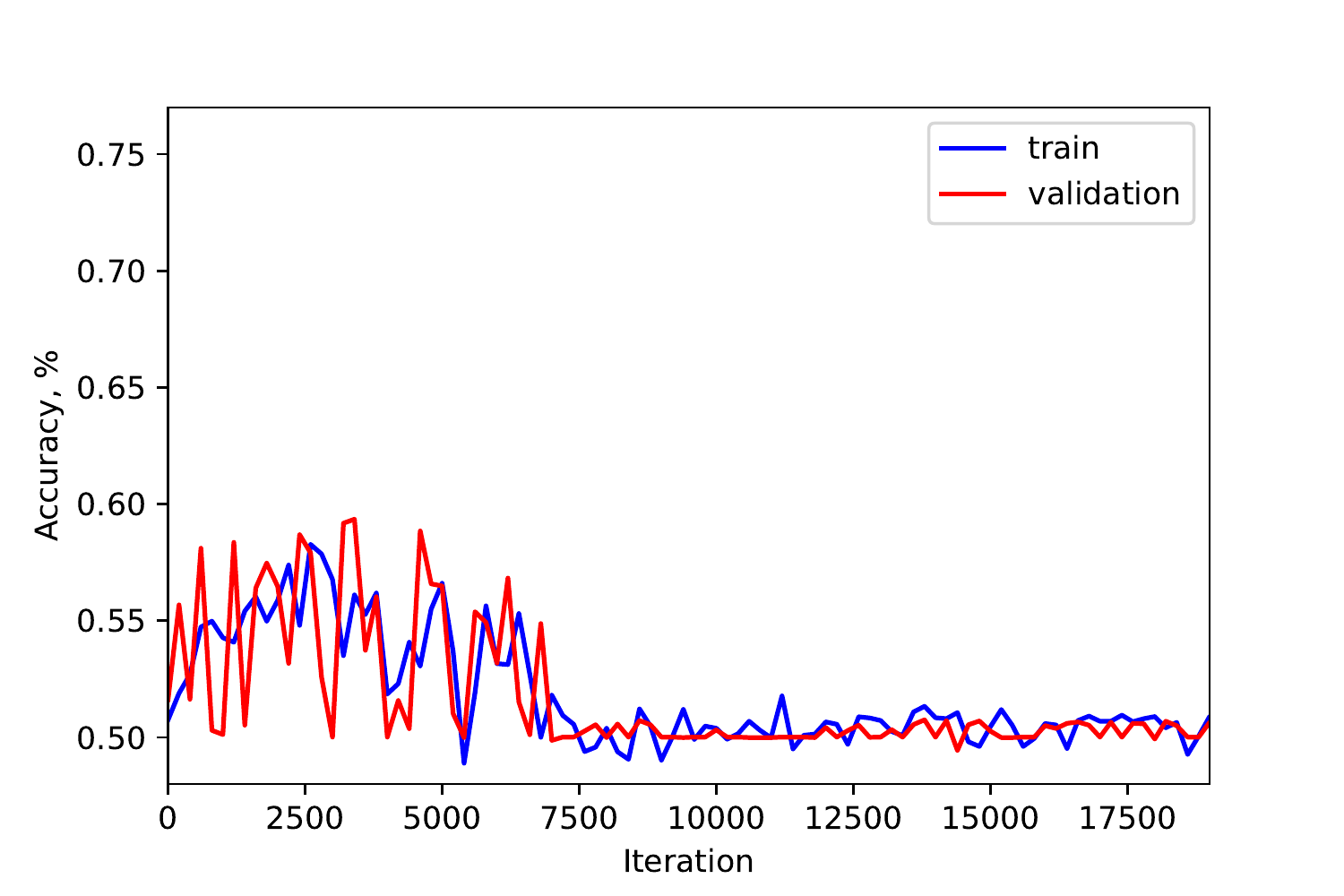}
         \caption{SimleAttention \\
         followed by a linear layer}
         \label{fig:with lin}
     \end{subfigure}
     \hfill
     \begin{subfigure}[b]{0.33\textwidth}
         \centering
         \includegraphics[width=\textwidth]{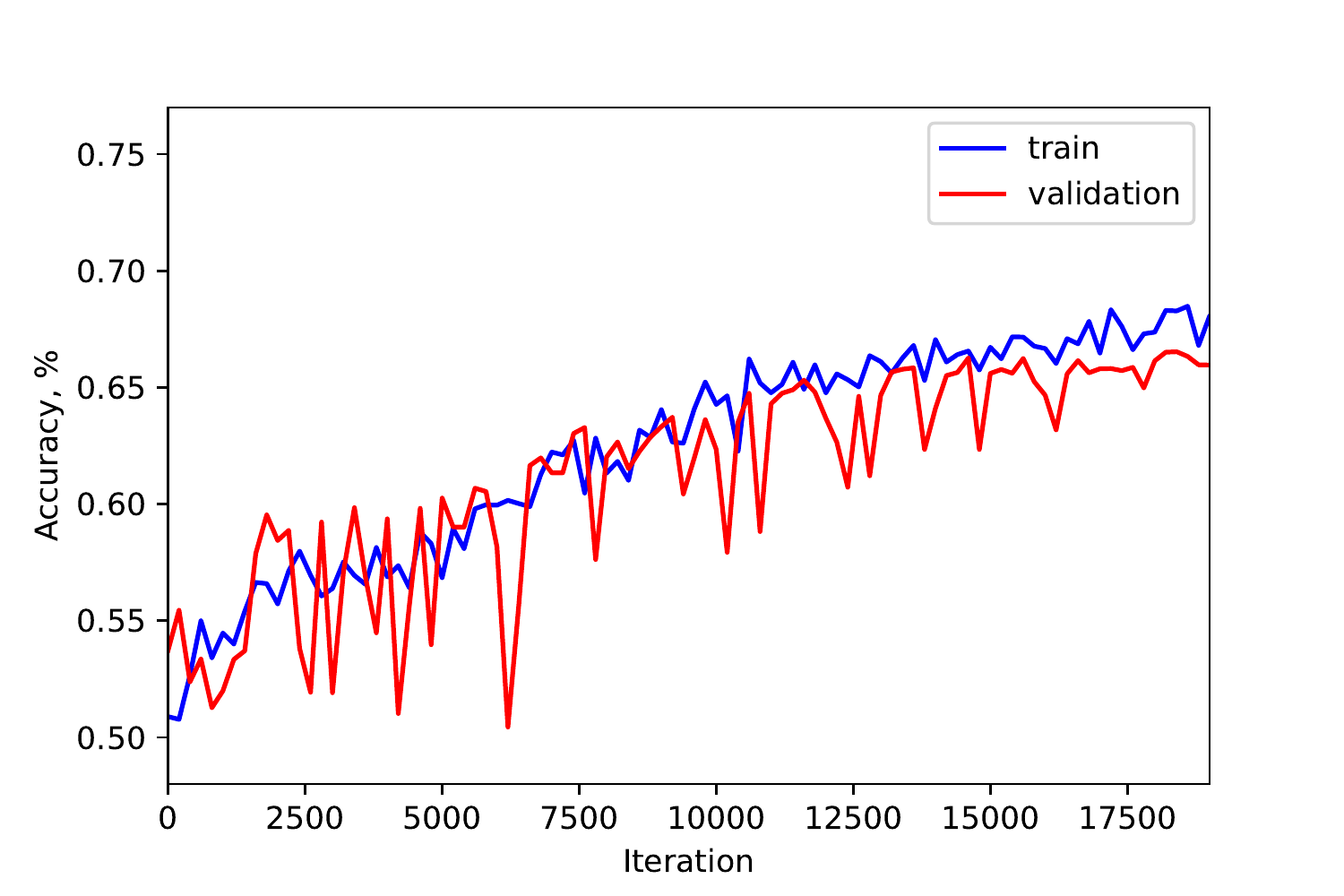}
         \caption{SimleAttention \\
         without a linear layer}
         \label{fig:no lin}
     \end{subfigure}
     \hfill
     \begin{subfigure}[b]{0.33\textwidth}
         \centering
         \includegraphics[width=\textwidth]{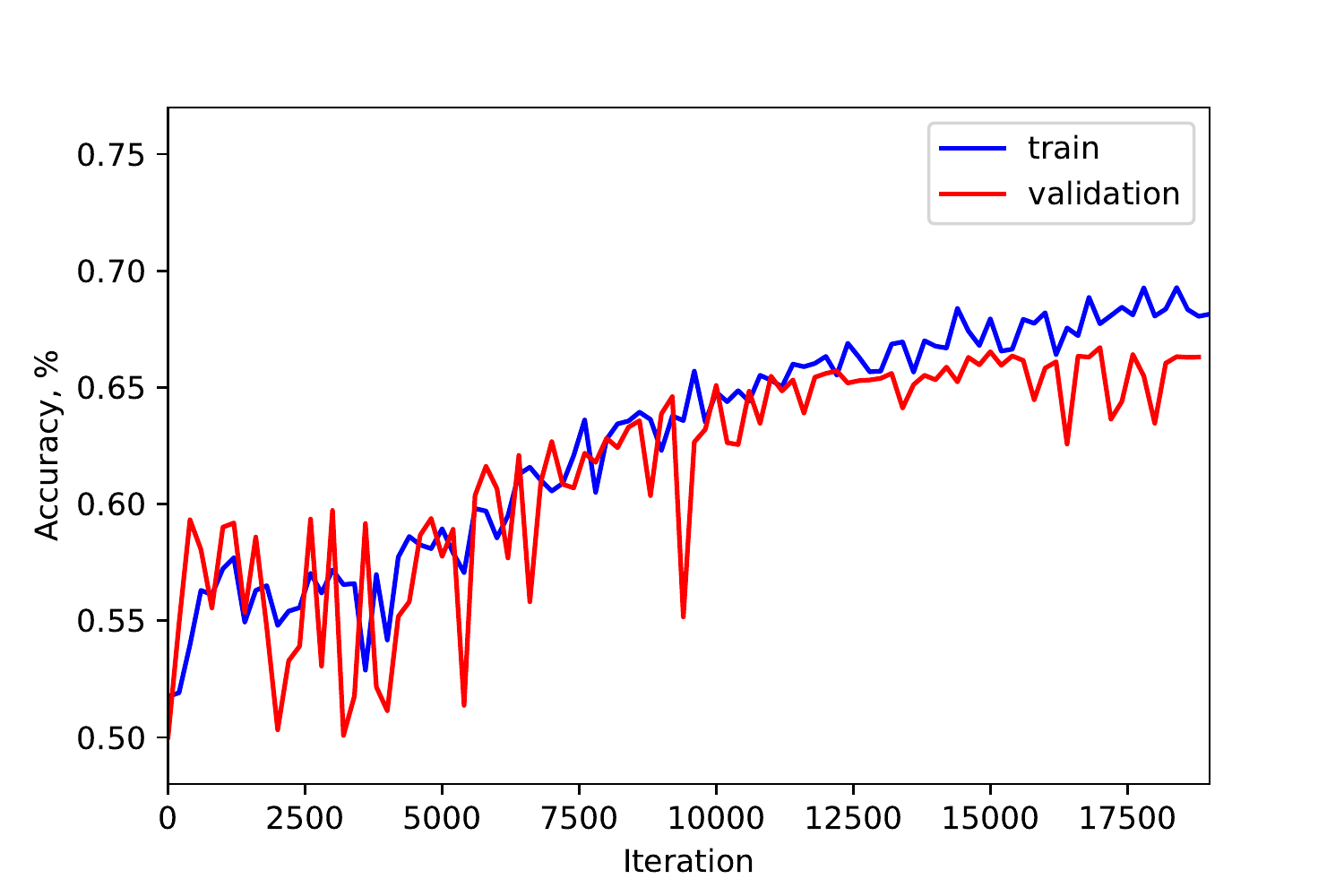}
         \caption{SimleAttention with a linear layer \\
         and skip connections}
         \label{fig:lin and skips}
     \end{subfigure}
        \caption{Training evolution plots for a SimpleAttention model containing 8 blocks, on text classification task.}
        \label{fig:add skips}
\end{figure}

\subsection{Larger models and utilizing weight from pretrained BERT}

As shown above, our architecture is superior on long text classification from the LRA benchmarks, which is unified test for efficient Transformer models. However, the true power of Transformer architecture is its ability to capture patterns from the large scale comprehensive datasets (often natural language datasets). Therefore, we performed preliminary experiment on comparing BERT \cite{devlin2019bert} language model with \emph{SimpleTRON} of the similar architecture.
Training from scratch of AG News Corpus dataset \cite{Zhang2015CharacterlevelCN}, showed that \emph{SimpleTRON} (with skip connections and linear layer) model with the architecture mimicking BERT posessed 89.9\% of accuracy, while training vanilla Transformer with BERT-base architecture we obtained 1.2\% higher accuracy.

However, while \emph{SimpleTRON} model in this experiment contained linear layer, the weights from BERT are fully transferable to the proposed architecture. Therefore, using weights from pretrained BERT model, we were able to perform fine-tuning on \emph{SimpleTRON} architecture. Interestingly, even though \emph{SimpleTRON} is in fact a different model we could obtain an inference gain using the weights from pretrained model, with the accuracy of 92.7\%, which was, however, still  1.5\% lower than pre-trained BERT model fine tuned on AG News dataset.

As discussed above, \emph{SimpleTRON} architecture works especially well, when there are limited number of stacked blocks, therefore we performed the experiments on the models with reduced depth both for fine-tuning and training from scratch. Indeed, \emph{SimpleTRON} architecture was found to outperform BERT based model, when two models containing 6 blocks were trained from scratch. While transferring 6 blocks of pretrained BERT a notable increase of performance was observed, 6 blocks \emph{SimpleTRON} model lags only a small margin behind the original BERT architecture. Overall, we found that even-though \emph{SimpleTRON} blocks are able to outperform Transformer architecture, in case of larger models, the proposed architecture do not take advantage of a stacked blocks well. This is a subject for a further investigation of \emph{SimpleTRON} architecture training and regularization. The performances of the models on depth from 1 to 12 blocks show that testing accuracy of our model saturates quickly with depth and may even decrease, when vanilla using vanilla self-attention mechanism inference accuracy increases with the model depth.


\begin{table}[t]
\caption{Model's performance with respect to the number of layers in the model. Training on AG News Corpus dataset.}
\label{sample-table}
\vskip 0.15in
\begin{center}
\begin{small}
\begin{tabular}{lcccr}
\toprule
$N_{blocks}$ & BERT & \emph{SimpleTRON}\\
\midrule
1    & 89.12 & 90.90 \\
2    & 90.38 & 90.32 \\
3    & 90.28 & 90.44 \\
4    & 90.66 & 90.41 \\
5    & 90.11 & 90.35 \\
6    & 90.30 & 90.22 \\
7    & 90.68 & 90.19 \\
8    & 90.68 & 89.98 \\
9    &  91.13 & 89.95 \\
10    & 91.21  & 89.94 \\
11    & 90.78 & 89.43 \\
12    & 90.10 & 89.90 \\
\hline
6$^*$    & 92.70 & 92.30 \\
12$^*$    & 94.20 & 92.70 \\
\bottomrule
\end{tabular}
\begin{tablenotes}
    $^*$ fine-tuning
\end{tablenotes}
\end{small}
\end{center}
\vskip -0.1in
\end{table}

\section{Conclusion and Future Work}

To conclude, we raised the question, whether the attention or any of its approximation is needed for the model performance. We presented a simple alternative of a truly linear model with a respect to the input length, that outperformed existing models on the several LRA tasks, at the same time possessing an extremely fast and memory efficient training. The key point is to reject the $\mat Q_h \mat K_h^T$ product with the following Softmax normalization and the linear layer after. We showed that trained q-k-v in the \emph{SimpleAttention} model can be effectively transferred to the classical \emph{SoftmaxAttention}, which allows fast model pre-training. Moreover as layers in the model are identical to the ones of transformer, weight transfer from pretrained large language models, such as BERT is possible with clear evidence of positive effect on the model performance. This is a valuable feature since training of large language models is a very resource demanding task. 
Nevertheless, there are several tasks yet to be done:
\paragraph{Different tasks} Transformer architecture is known to be pervasive, however we are fully aware that performance of any model can be task-dependent. Here we show the results on text-classification task from LRA benchmark dataset and AG News. Therefore thus our goal in the near future is to expand \emph{SimpleTRON} application to other modalities, such as computer vision, as well as looking for a more efficient way to utilize the model depth.
\paragraph{Interpretability} Initially the attention mechanism allowed a human-comprehensible  visualization of relevancy scores in the sequences, nevertheless many approximation models lost this feature. Therefore, our goal is to gain it back using our model.
\paragraph{q-k-v framework elimination} In the present work we followed the original q-k-v framework in order to show the step further towards attetnionless transformer architecture. However, we believe that there is a more efficient framework as long as q-k-v initially assumed global pairwise comparison.

\section{Acknowledgements:}

This research is supported by the Czech Ministry of Education, 
Youth and Sports from the Czech Operational Programme Research, 
Development, and Education, under grant agreement 
No. CZ.02.1.01/0.0/0.0/15003/0000421 and the Grant Agency of the Czech Technical University in Prague (SGS20/213/OHK3/3T/18)

\printbibliography
\newpage

\begin{appendices}

\section{Hyperparameters}

\newcommand{\tabletwowidth}{1.75cm}

\begin{table}[H]
    \centering
    \begin{tabular}{l | C{\tabletwowidth} C{\tabletwowidth} C{\tabletwowidth}}
        \hline
         & & & \\
         \textbf{Parameter} & \textbf{Classification} & \textbf{Matching} & \textbf{ListOps} \\
         & & & \\
         \hline
         Seq. Length & 4000 & 4000 & 2000\\
         Batch Size & 32 & 32 & 32\\
         Training Steps & 20 000 & 15 000 & 15 000\\
         Optimizer & \multicolumn{3}{c}{AdamW ($\beta_1 = 0.9$, $\beta_2 = 0.999$)}\\
         Base LR & 0.05 & 0.05 & 0.005\\
         Weight Decay & 0.1 & 0.1 & 0.1\\
         Warmup Steps & 8000 & 8000 & 1000\\
         Schedule & \multicolumn{3}{c}{Base LR * Warmup * Sqrt Decay}\\
         Warmup Mul. & \multicolumn{3}{c}{$min(1, \text{Current Step} / \text{Warmup Steps})$}\\
         Sqrt Decay Mul. & \multicolumn{3}{c}{$1 / \sqrt{max(\text{Current Step}, \text{Warmup Steps})}$}\\
         Loss & \multicolumn{3}{c}{CCE}\\
         Blocks & 4 & 4 & 6\\
         Heads & 4 & 4 & 8\\
         Hidden dim. & 256 & 128 & 512\\
         QKV dim. & 256 & 128 & 512\\
         MLP dim. & 1024 & 512 & 2048\\
         Dropout & 0.1 & 0.1 & 0.1\\
         Activation & GELU & GELU (ReLU in output) & GELU\\
         Pooling & CLS & CLS & CLS\\
         Pos. encoding & Learnable & Learnable & Learnable\\
         \hline
    \end{tabular}
    \caption{Hyperparameters used for this experiment}
    \label{tab:hyper}
\end{table}

\section{Parameters' Training Evolution}

\begin{figure}[h]
     \centering
     \begin{subfigure}[b]{0.48\textwidth}
         \centering
         \includegraphics[width=\textwidth]{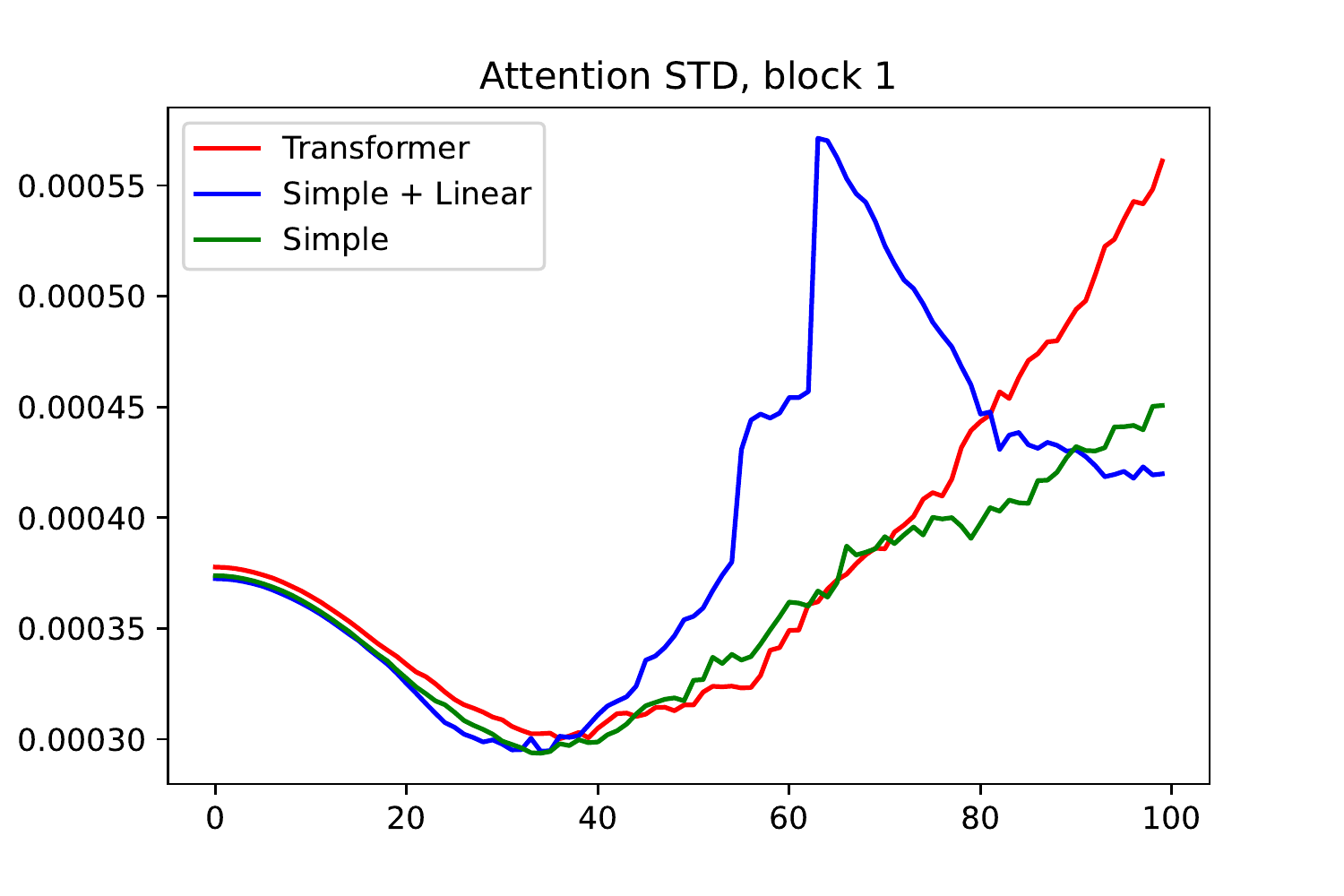}
         \caption{Block 1}
         \label{fig:with lin}
     \end{subfigure}
     \hfill
     \begin{subfigure}[b]{0.48\textwidth}
         \centering
         \includegraphics[width=\textwidth]{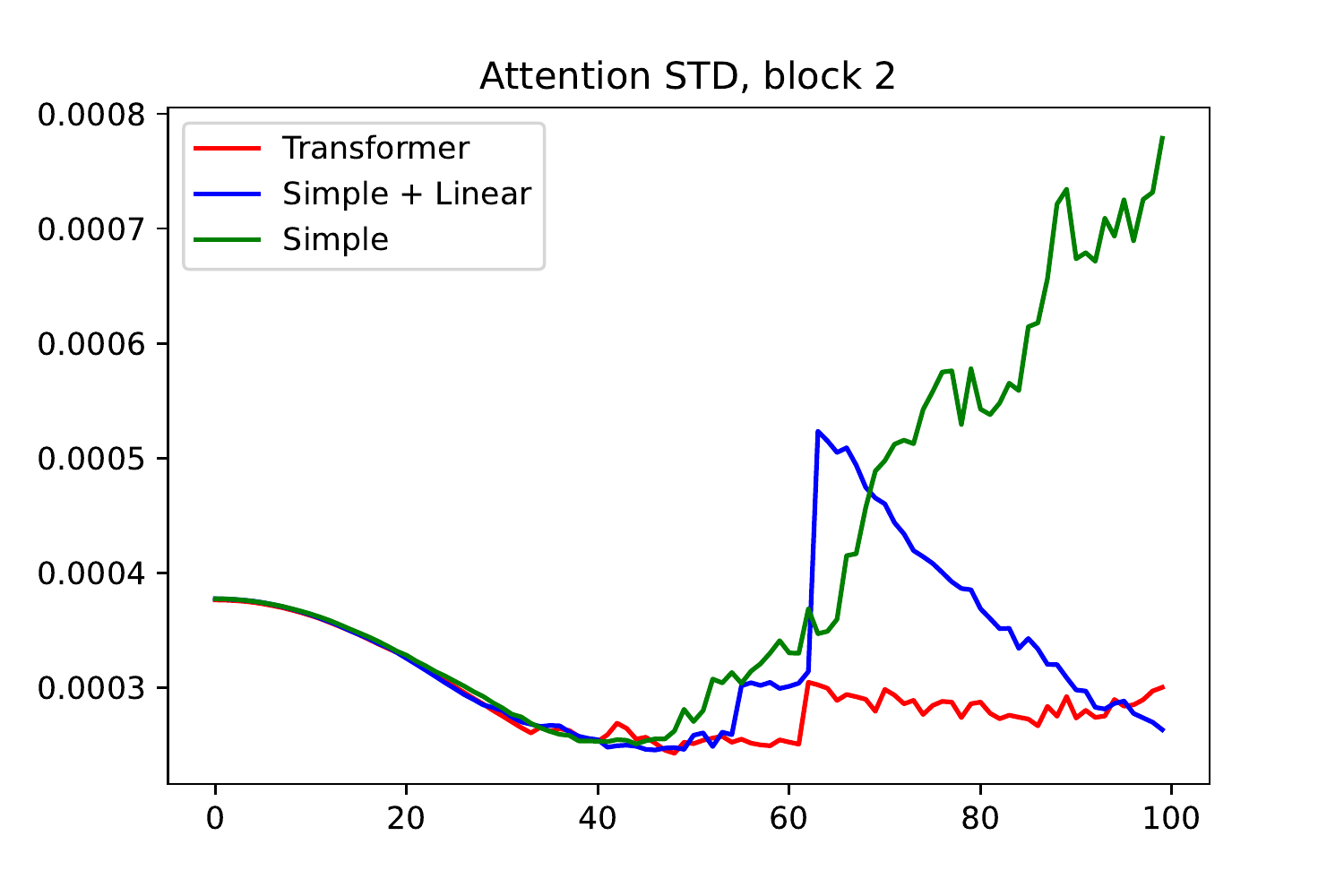}
         \caption{Block 2}
         \label{fig:no lin}
     \end{subfigure}
     \hfill
     \begin{subfigure}[b]{0.48\textwidth}
         \centering
         \includegraphics[width=\textwidth]{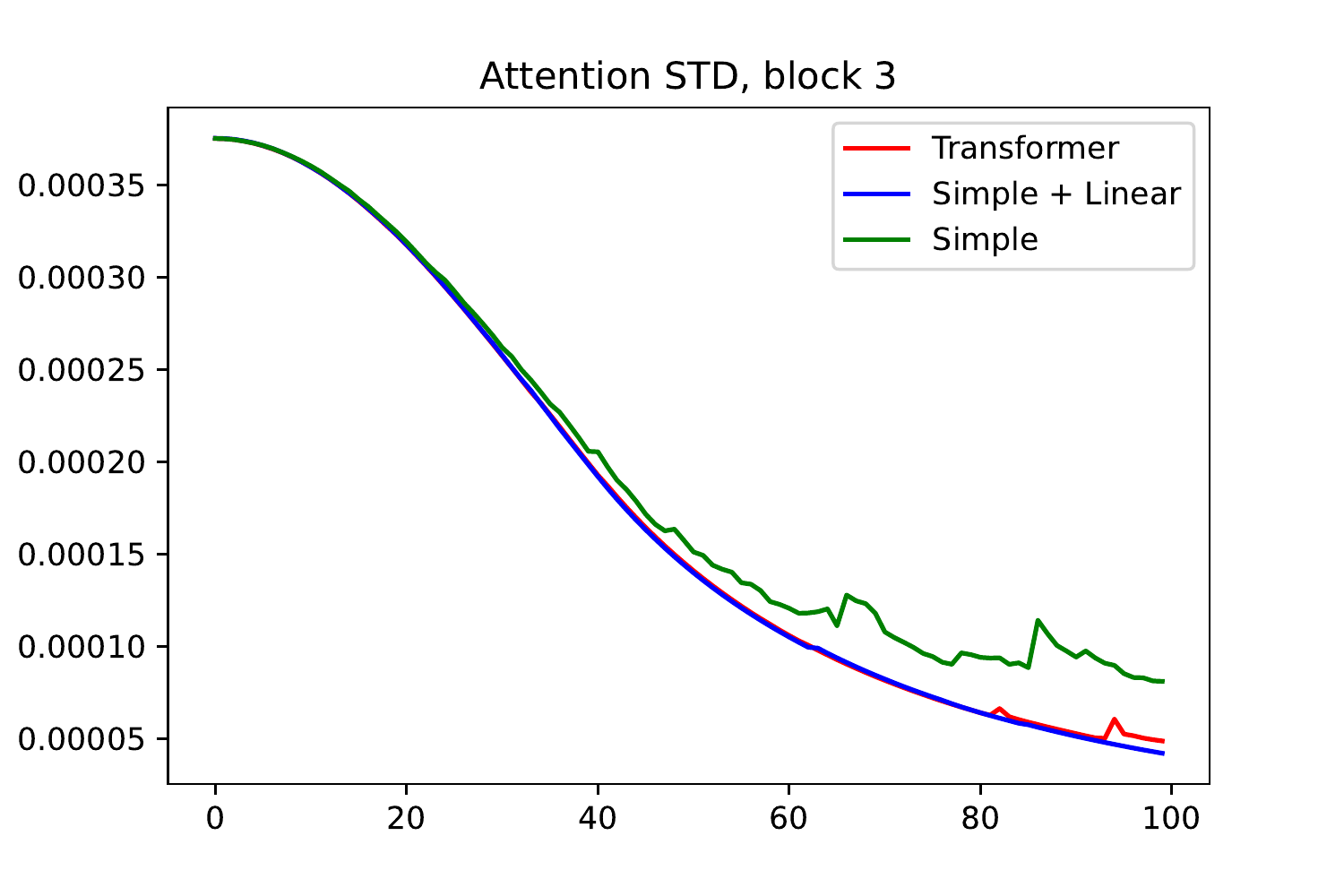}
         \caption{Block 3}
         \label{fig:lin and skips}
     \end{subfigure}
     \begin{subfigure}[b]{0.48\textwidth}
         \centering
         \includegraphics[width=\textwidth]{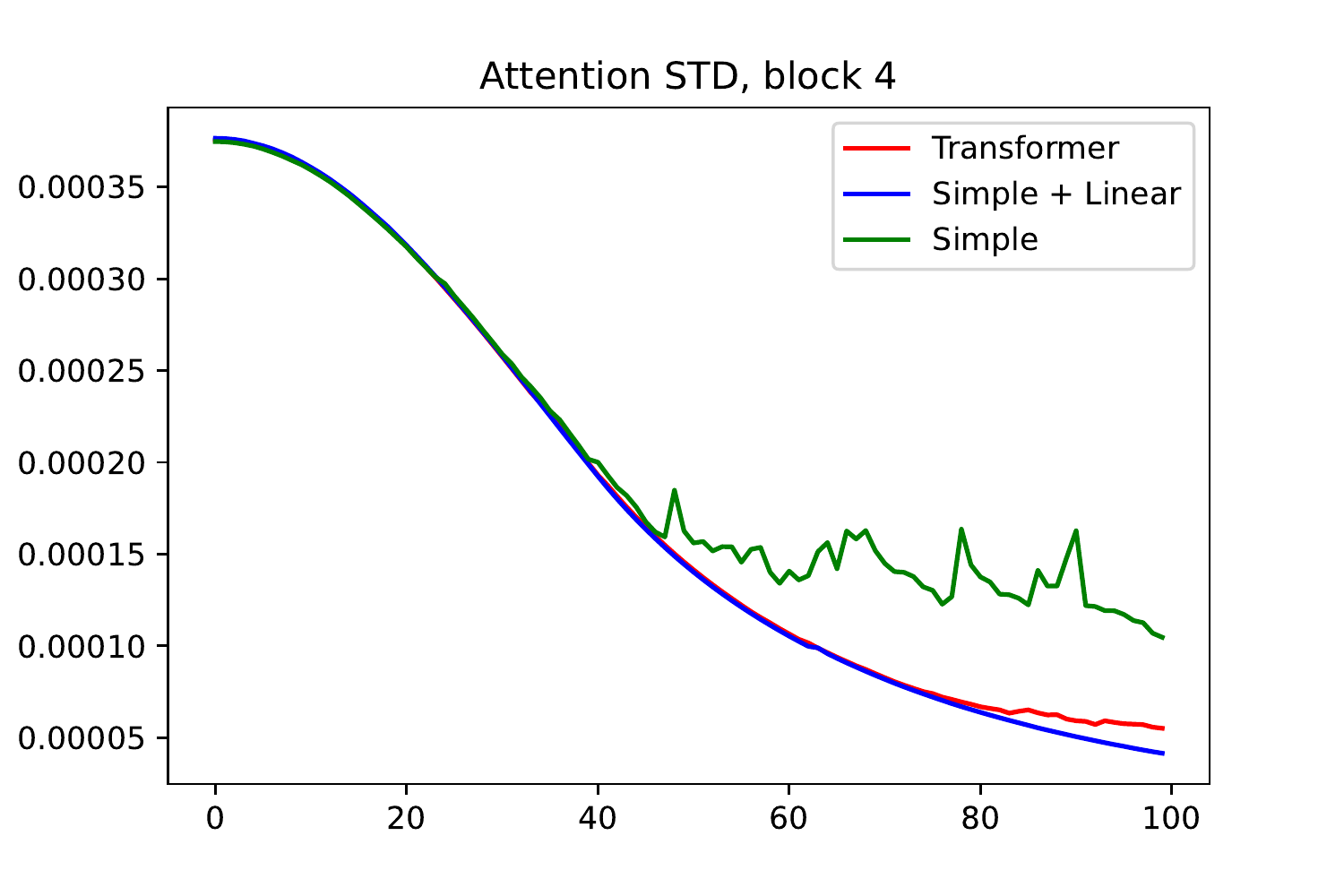}
         \caption{Block 4}
         \label{fig:lin and skips}
     \end{subfigure}
        \caption{Training evolution of standard deviation of Attention output weights for vanilla transformer ($\text{softmax}(\frac{QK^T}{\sqrt{d}})V$) and SimpleTRON ($\frac{1}{\sqrt{L}}QK^TV$) model containing 8 blocks, on text classification task. In case of vanilla transformer Softmax normalization is omitted in standard deviation calculation.}
        \label{fig:attention weight std}
\end{figure}

\end{appendices}

\end{document}